# LLM Predictive Scoring and Validation: Inferring Experience Ratings from Unstructured Text

Andrew Hong, Jason Potteiger, Ito Zapata

*Dimension Labs*

April 2026

**Abstract**

We tasked GPT-4.1 to read what baseball fans wrote about their game-day experience and predict the overall experience rating each fan gave on a 0–10 survey scale. The model received only the text of a single open-ended response. These AI predictions were compared with the actual experience ratings captured by the survey instrument across approximately 10,000 fan responses from five Major League Baseball teams. In total two-thirds of predicted ratings fell within one point of self-reported fan ratings (67% within ±1, 36% exact match), and the predicted measurement was near-deterministic across three independent scoring runs (87% exact agreement, 99.9% within ±1). Predicted ratings aligned most strongly with the overall experience rating (r = 0.82) rather than with any specific aspect of the game-day experience such as parking, concessions, staff, etc. However, predictions were systematically lower than self-reported ratings by approximately one point, and this gap was not driven by any single aspect. Rather, our analysis shows that self-reported ratings capture the fan's verdict: an overall evaluative judgment that integrates the entire experience. While predicted ratings quantify the impact of salient moments characterized as memorable, emotionally intense, unusual, or actionable. Each measure contains information the other misses. These baseline results establish that a simple, unoptimized prompt can directionally predict how fans rate their experience from the text a fan wrote and that a gap between the two numbers can be interpreted as a construct difference worth preserving rather than an error to eliminate.

# 1. Introduction

A fan walks out of a ballpark, opens a survey on her phone, and rates her overall experience a 10. Then she writes about what happened: the concession lines were brutal, parking cost forty dollars, and the team lost. A language model reads her words and predicts a 5. Both numbers are accurate descriptions of something, but they are not describing the same thing. The survey score is the fan's explicit verdict[1] on the day: a single number that integrates the highs, the lows, and everything she did not bother to mention (including some aspects of team affinity). The text captures what is most salient[2], what she chose to talk about, shaped by whatever was most vivid, most frustrating, or most worth reporting. These are related but separable constructs, and the distance between them contains information that neither captures alone. This paper asks whether a large language model (LLM) can reliably recover the score a fan gave from the text a fan wrote, and what it means when the two diverge.

We used GPT-4.1 to predict overall experience ratings from the open-ended survey responses of 9,977 fans who attended games across five MLB teams in the first several months of the 2025 season. The model received only the fan's written text, a single open-ended survey response in which the fan described their game-day experience in their own words. It received no survey scores, no aspect ratings, no demographic information, and no knowledge of the venue, the team's record, or the outcome of the game. The prediction task was deliberately minimal: a simple prompt asked the model to estimate the rating the fan would give on a 0–10 scale, using only evidence in the text. We ran the prediction three times independently to assess whether the measurement was reproducible or sensitive to the stochasticity (randomness or unpredictability) inherent in language model inference. The prompt was not optimized for this domain and included no calibration, no examples, and no instructions for handling mixed sentiment. This design was intentional: we wanted to establish baseline performance for the simplest viable approach before introducing complexity.

A rating, a single number on an 11-point scale, tells an organization the direction of satisfaction but not the salient reason behind it. The open-ended text can. Analyzing this data promises specific, actionable information that the rating compresses into a single digit. But unstructured textual data cannot be read by humans at scale. The practical choice is often between extracting meaning from text automatically or not extracting it at all. This study examines *what* an LLM extracts when it reads customer text, how reliably it does so, and what the relationship between that extraction and the fan's own rating reveals about the structure of experience measurement.

---

[1] Verdict: an explicit overall judgment that integrates their entire day, including the parts they did not mention.
[2] Salience: moments that were most memorable, emotionally intense, unusual, or actionable.

The results establish that for a majority of records feedback recovery of the self-reported rating is possible. Two-thirds of predicted ratings fell within one point of the fan's self-reported score, and just over one-third matched exactly. Even more impressive was repeatability. The predicted measurement was found to be near-deterministic: across three independent runs, 87% of predictions were identical and 99.9% were within one point. Predictions tracked daily trends in fan experience with a correlation of 0.92. Overall, the model tended to underpredict self-reported ratings by roughly one point on average, and this underprediction concentrated in the middle-to-high range of the scale. More precisely, among the fans who rated their day highly but wrote about what went wrong. These results raise two questions the second half of this paper addresses: (1) what does the systematic gap tell us about what text captures versus what surveys measure, and (2) is perfect alignment between predicted and self-reported ratings the right goal?

## 2. LLMs as measurement instruments and the constructs they recover

The use of large language models to quantify attributes in unstructured text has moved rapidly from proof-of-concept to methodological framework. Asirvatham, Mokski, and Shleifer (2026) present the most comprehensive validation to date, demonstrating through their GABRIEL system that GPT performs at levels generally indistinguishable from human evaluators across more than 1,000 annotation tasks spanning congressional remarks, social media content, and educational curricula. Their finding that labeling results do not depend on the exact prompting strategy suggests that when a construct is clearly defined, the model's measurement is robust to minor prompt variation: a finding directly relevant to the operational deployment we examine here.

This robustness, however, should not be mistaken for unconditional reliability. Pangakis, Wolken, and Fasching (2023) replicate 27 annotation tasks across 11 non-public datasets drawn from high-impact social science journals and find that performance is highly contingent on the dataset and task type: nine of 27 tasks had either precision or recall below 0.5, despite a median accuracy of 0.85. The aggregate number looks strong; the task-level variation does not. Their central argument is that any automated annotation process must validate against human-labeled data on a task-by-task basis, because the factors that degrade performance (prompt quality, text idiosyncrasies, conceptual difficulty) persist even as the underlying models improve. The validation-first design of this study follows directly from this principle. We validate before interpreting.

These studies address constructs that are categorical: a speech has a topic, a tweet expresses a stance, a survey response does or does not mention a theme. The construct we measure here — overall experience on a 0–10 scale — is not categorical. It is scalar, and scalar measurement with LLMs introduces problems that do not arise in classification. Consider what it means to ask a model to assign a 6 rather than a 7. Is that distinction meaningful? Are the intervals equal — is the distance from 3 to 4 the same as from 8 to 9? Does the model use the full range of the scale, or does it compress into a subset? Licht, Sarkar, Wu, Goel, Stoehr, Ash, and Hoyle (2025) demonstrate that these are not hypothetical concerns. When prompted to score texts on a numeric scale, LLMs produce distributions that exhibit heaping (probability mass concentrated on a few values rather than spread across the scale) and systematic compression, where the model avoids the extremes the scale permits. Different models produce different distributions for the same texts, and inter-model agreement can be low. These are properties of the measurement instrument, not of the texts being measured.

Licht et al. evaluate four approaches to scalar measurement with LLMs: unweighted direct pointwise scoring, token-probability-weighted pointwise scoring, aggregation of pairwise comparisons, and fine-tuning. They find that token-probability-weighted averaging partially corrects scale distortion, but the smoothing effect diminishes with larger models whose confidence in the modal response is so high that the weighted average collapses back toward it. Pairwise comparisons, asking the model which of two texts reflects a higher score rather than assigning an absolute number, show greater robustness to prompt variation and model drift. Our study uses unweighted pointwise scoring, the simplest of the four approaches, which means the compression and scale distortion Licht et al. document are expected features of our measurement. This design choice was deliberate: we wanted a baseline. But it also means we cannot fully separate construct-level underprediction (the text genuinely emphasizes different things than the rating captures) from method-level compression (pointwise scoring systematically avoids scale extremes). Future work should test whether pairwise or probability-weighted approaches reduce the magnitude of underprediction, which would isolate how much of the gap is measurement artifact versus construct difference.

These three bodies of work share an implicit assumption: the construct the model measures is the construct the researcher intends. Halterman and Keith (2024) challenge this assumption directly, finding that when codebook definitions deviate from LLM pretraining priors, zero-shot classification may recover background concepts absorbed during training rather than the researcher-specified construct. When an LLM rates how "pro-innovation" a speech is, or classifies the stance of a tweet, the target construct is researcher-defined and the text is the object of measurement. Our setting introduces a different complication. We are not measuring an attribute of the text itself. We are asking the model to infer a number that lives in the respondent's head, the overall experience rating they would (and did) give on a survey, from text that may or may not reflect that number faithfully. The text is a window, not a mirror. This means that disagreement between the model's prediction and the respondent's rating could reflect model error, but it could also reflect a genuine difference between what people write about and what they report.This aspect of the survey instrument deserves emphasis because it eliminates the most obvious alternative explanations for divergence. In the survey used here, the overall experience rating (0–10) is followed immediately by the open-ended prompt: "You selected [rating] to describe your overall experience. Please tell us why you selected that number." The fan has just committed to a number and is explicitly asked to explain it. The same events are equally accessible. There is no temporal decay, no context shift, no differential recall. And yet we find examples where a fan rates her experience a 10 and then, asked to explain that 10, writes about long concession lines and expensive parking. Rather than being inconsistent she is performing two different cognitive operations in sequence: the rating is a holistic evaluative judgment; the explanation a retrieval of salient, reportable incidents. The survey design makes this construct difference maximally visible because it holds everything else constant. If the text diverges from the rating even when the respondent is explicitly asked to connect them, the divergence is structural, not artifactual.

Schwarz (1999) demonstrates that question format determines which cognitive operation a respondent performs. Open-ended prompts activate what he terms accessibility effects: respondents retrieve specific, vivid, and recently salient incidents and report on those. Rating scales activate a different process: holistic evaluative judgment that draws on global affective summary as opposed to specific incident recall. Schwarz and Clore (1983) formalize this as the mood-as-information model: when asked for an overall evaluation, respondents use their current affective state as a direct input rather than summing incidents, which means routine positives that don't generate strong affect go unrepresented even in an honest answer. The result is that a fan completing the same survey in the same moment can truthfully report a 10 and write about the parking, because the rating scale is asking for one thing and the open-ended prompt is asking for another.

This body of work supports the hypothesis that the gap between text-derived predictions and survey scores is not a measurement artifact but a construct difference built into the instruments themselves. The validation literature largely grounds itself in the assumption that human-coded labels are a gold standard. However, this assumption is empirically fragile. Törnberg (2023) argues that disagreements between LLM and human coders should not be understood as fine-tuning to match human data, since human coding cannot be treated as an unquestioned gold standard; instead, such disagreements represent a process of mutual learning through which rigorous definitions of measured constructs are iteratively refined. Research conducted by Glazier, Boydstun, and Feezell (2021) provide direct empirical evidence. They asked survey respondents to categorize their own open-ended answers and compared those self-codes to labels assigned by trained researchers using a standard codebook. Agreement was modest (Cohen's $\kappa = 0.46$), and the divergence was not random: for identical text, respondents and coders systematically assigned different meanings. If human coders cannot reliably reconstruct what respondents meant from the text those respondents wrote, the expectation that an LLM should match self-reported labels exactly is equally unreasonable. This study treats the fan's self-reported rating as a reference point for understanding construct divergence, not as a truth against which predictions are scored. We return to this interpretation in Section 8, after establishing the empirical facts of alignment and divergence.

## 3. Data, annotation, and validation design

### 3.1 Data source and sample

The dataset consists of post-game survey responses collected from fans of five Major League Baseball teams in the spring of 2025 between March 28 and June 23. Fans received a survey following their game-day visit. The survey included a closed-ended overall experience rating (`overall_numrat`: "On a scale from 0 to 10, how would you rate your overall experience at the [club] game on [date]?") and an open-ended question inviting the fan to explain the reason for their rating (...Please tell us why you selected that number), followed by a battery of specific experience aspect ratings.

After restricting to sessions with valid overall ratings (numeric, 0–10) and non-null text responses, the working sample contains 9,977 fan sessions. The mean self-reported overall experience rating is 7.68 (SD ≈ 2.4), with a distribution skewed toward the high end: 49% of fans rated their experience 9 or 10, and 7% rated it 0–2. The median open-ended response was approximately 351 characters (interquartile range: 283–463).

### 3.2 Annotation methodology

We used GPT-4.1 to predict overall experience ratings from the fan's open-ended text. The model operated as a recognition task: it read the text and inferred a structured output based on patterns in the language. The predicted rating field is a scalar measurement task: the model places each response on a continuous numeric scale rather than assigning a categorical label, which introduces measurement properties (compression, calibration, scale usage) that are specific to scalar constructs (Licht et al., 2025). It did not receive the fan's survey scores, aspect ratings, or any other contextual information.

The annotation was executed on the Dimension Labs language data platform using a prompt function, a JSON schema that defines the model's task and output structure. This method produced the three fields described in Table 1. Each fan's textual response was processed as individual sessions: the model received one session's open-ended text as input and returned a structured set of labeled fields for that session alone. It had no access to other fans' responses and no memory across sessions. Session-level independence is a design feature, not a limitation: it ensures that each prediction reflects only the language in a single response, making the resulting ratings comparable across fans, teams, venues, and time periods without the confound of cross-session information.

*Table 1. Attributes measured in the validation study, with source and operational definition.*

| Attribute | Source | Definition |
|---|---|---|
| `predicted_rating` | LLM-annotated (GPT-4.1) | Estimated overall experience rating (0–10, integer). Prompt: "Estimate the fan's self-reported overall experience rating as they would answer the survey question. Use ONLY evidence in the fan's text." |
| `predicted_rating_confidence_score` | LLM-annotated (GPT-4.1) | Evidence quality: High (explicit rating or clear sentiment), Medium (mixed/limited cues), Low (ambiguous). |
| `predicted_rating_evidence` | LLM-annotated (GPT-4.1) | In fifteen words or fewer, the strongest textual evidence supporting the inferred rating. |
| `overall_numrat` | Self-reported (fan survey) | "On a scale from 0 to 10, how would you rate your overall experience at the [team] game on [date]?" |

*Full prompt schema appendix A*

The prompt was deliberately simple and unoptimized. It asked the model to infer the survey score the fan would give. It did not instruct the model to weigh positive and negative signals in any particular way, did not provide examples, and did not calibrate for known biases. This design was intentional: we wanted to establish baseline performance for the simplest viable prompt before introducing complexity.

### 3.3 Validation design

We ran the prompt function three times independently on the same text data, producing three complete sets of predictions (runs p2_1, p2_2, and p2_3). Each run processed all sessions. Null predictions (where the model judged insufficient evidence) occurred in 20–25 sessions per run, less than 0.3% of the sample.

We evaluate alignment using four metrics. **Exact match rate** is the proportion of predictions equal to the fan's score. **Within ±1 agreement** is the proportion where |predicted − actual| ≤ 1; on an 11-point scale, this is the primary metric for operationally useful accuracy. **Mean absolute error (MAE)** is the average magnitude of prediction error. **Bias** is the mean signed difference (predicted − actual). We also assess rank-order agreement (Spearman correlation), calibration, consistency across score bands, and the diagnostic value of the model's confidence labels.

## 4. Predicted ratings align with fan-reported overall experience

Consider two ways a measurement can fail. It can be wrong (predicting a 3 when the fan said 9). Or it can be noisy (predicting a 7 on Tuesday and a 4 on Wednesday from the same text). The first is an accuracy question. The second is a reliability question. This section addresses accuracy. The next addresses reliability.

Across approximately 10,000 evaluated sessions, two-thirds (67%) of predicted ratings fell within one point of the fan's self-reported score, and over a third (36%) matched exactly. The model's primary systematic error was underprediction: on average, the predicted rating was about one point lower than the fan's score.

*Table 2. Core alignment between predicted and self-reported overall experience ratings.*

|      | N      | Exact match | Within ±1 | MAE  | Bias  |
|------|--------|-------------|-----------|------|-------|
| P2_1 | 10,428 | 36.4%       | 67.5%     | 1.29 | −0.96 |
| P2_2 | 10,433 | 36.2%       | 67.4%     | 1.29 | −0.96 |
| P2_3 | 10,425 | 36.4%       | 67.3%     | 1.29 | −0.96 |

*95% margins of error for all agreement rates are ±0.9 percentage points.*

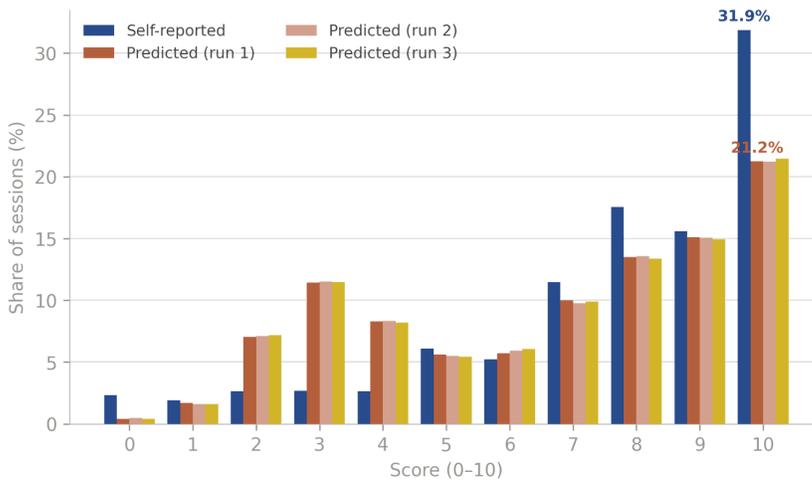

*Figure 1. Predicted scores compress the distribution downward. Three independent runs produce nearly identical predicted distributions, all shifted from the 9–10 range into the 2–4 range.*

The measurement is stable at the aggregate level, and Section 5 will show it is stable at the session level as well.

To put the accuracy number in more context, within ±1 agreement means that on an 11-point scale the prediction falls in a 3-point window centered on the fan's score (i.e., a self-reported rating of 7 receives a prediction of 6, 7 or 8). With no prior information about the fan, the venue, or the game, as well as no prompt engineering, this represents a strong baseline. Additionally, the rank-order agreement between predicted and actual scores (Spearman $\rho = 0.86$) indicates that the model preserves the relative ordering of fan experiences with high fidelity: fans who reported better experiences received higher predicted scores, and vice versa.

**Alignment is strongest at the top and bottom of the scale.** The score-band breakdown reveals that the model performs best where fan language is most unambiguous. At the high end (actual 9–10), 83% of predictions fell within one point and over half matched exactly. At the low end (actual 0–2), 63% were within one point. The hardest region is the broad middle: fans who rated their experience 7–8 saw only 54% within ±1 agreement, with a mean error of nearly two points.

*Table 3. Alignment metrics by self-reported score band, showing concentrated underprediction in the 7–8 range.*

| Score band | N | Exact match | Within ±1 | MAE | Bias |
| --- | --- | --- | --- | --- | --- |
| Low (0–2) | 714 | 22.0% | 62.9% | 1.34 | +1.20 |
| Moderate (3–6) | 1,736 | 18.7% | 49.4% | 1.61 | −1.23 |
| High (7–8) | 3,027 | 25.2% | 53.5% | 1.86 | −1.64 |
| Very high (9–10) | 4,951 | 51.6% | 83.2% | 0.81 | −0.76 |

The bias column tells the story. At the low end, the model overpredicts by 1.2 points — it is more generous than fans who rated poorly. At the high end, it underpredicts by 0.76 points. In the critical 7–8 band, underprediction reaches 1.64 points. This is where the "happy but..." pattern lives: fans who had a good day but wrote about what went wrong. The model reads the complaint; the fan reports the verdict.

Two examples ground the extremes. A fan who wrote "Every team member encountered was pleasant and interactive, seats were great, ballpark is beautiful" rated her experience a 10; the model predicted 10. A fan who wrote about waiting forty minutes at concessions and "barely saw any of the game" rated his experience a 1; the model predicted 2. Where the text and the verdict point in the same direction, the model follows. Section 7 examines the more informative cases where they do not.

**The calibration curve confirms a compressed scale.** When the model predicts a 10, the average fan score is 9.9 — excellent calibration. But when it predicts a 6, the average fan score is 7.4. The predicted scale is compressed and more critical relative to the survey scale: a predicted 6 roughly corresponds to a fan who actually rated the experience a 7, not a fan on the fence. This compression is consistent with the systematic scale distortion Licht et al. (2025) document in LLM pointwise scoring, where models avoid the extremes of the scale and concentrate predictions in the mid-range. This compression is systematic and monotonic, which means it is correctable through post-hoc calibration if numeric comparability with survey scores is desired.

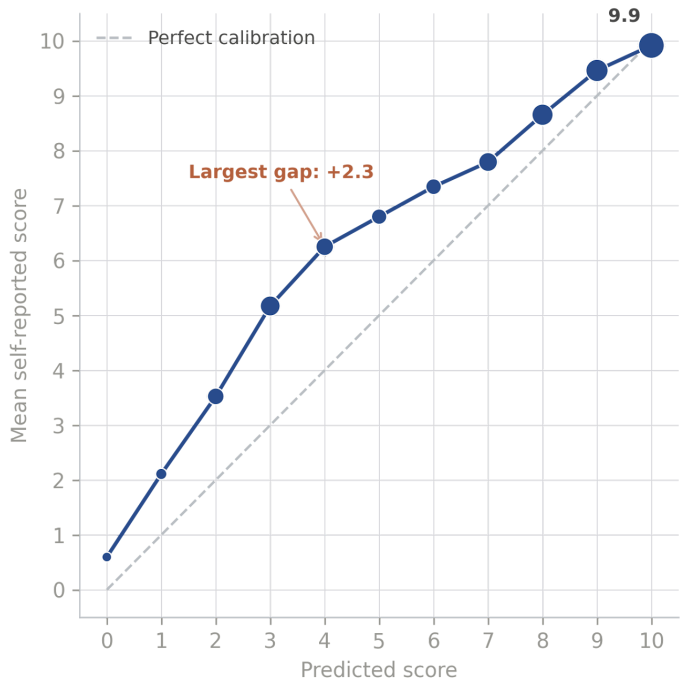

Figure 2. The calibration curve is monotonic but systematically above the 45° reference line, confirming that predicted scores are compressed relative to the survey scale. The gap peaks in the mid-range where text mixes positive and negative signals.

**Confidence labels are highly diagnostic.** The prompt function produced a confidence score alongside each prediction (High or Medium; Low was not observed in the evaluated sample). This label transforms the measurement's utility. Among High-confidence predictions, within ±1 agreement rises from 67% to 82%, exact match doubles from 36% to 50%, and mean absolute error drops from 1.29 to 0.87. Among Medium-confidence predictions, within ±1 falls to 47% and MAE rises to 1.88.

Table 4. Confidence labels are strongly diagnostic: High-confidence predictions are substantially more accurate.

| Confidence | N | Exact match | Within ±1 | MAE | Bias |
|---|---|---|---|---|---|
| High | 6,166 | 50.4% | 81.5% | 0.87 | −0.55 |
| Medium | 4,262 | 16.1% | 47.3% | 1.88 | −1.56 |

What is the confidence label actually measuring? It is not a report of the model's internal certainty at the moment of prediction. It is a judgment about the coherence between the text and the rating the model assigned that describes how clearly the text signals an overall experience level. High confidence means the text contained explicit ratings, strongly directional sentiment, or unambiguous overall evaluations. Medium confidence means the text was mixed, limited, or contained conflicting signals. In practice, the confidence label is a measure of **textual clarity**: how well the text alone can determine an overall experience score.

Two cases illustrate the distinction. A fan who wrote "Every team member encountered was pleasant and interactive, seats were great, ballpark is beautiful" received a High-confidence prediction of 10 against an actual score of 10. The text points unambiguously in one direction; the model follows it there. A fan who wrote "Everything was great… but parking… $40… rip off… will only attend if invited or win tickets" received a Medium-confidence prediction of 5 against an actual score of 10. The text contains a genuine conflict, a positive verdict and a negative grievance coexisting, and the model correctly flags that the signal is ambiguous. Medium confidence is not a failure mode. It is the model reporting that the text does not clearly resolve to a single overall score.

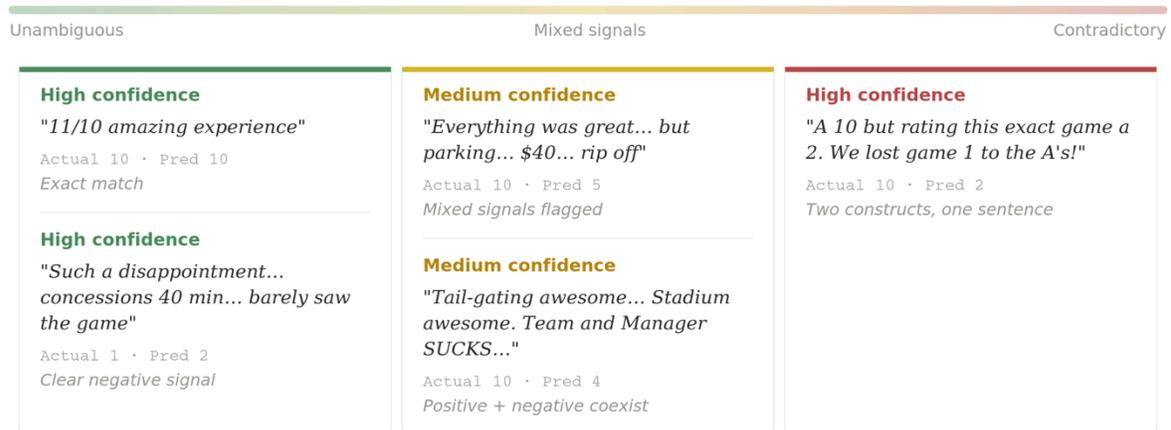

*Figure 3. Textual clarity, not confidence alone, determines prediction accuracy. High confidence means the text is explicit — but the rightmost case is High confidence and the largest gap, because the text is explicit about two different things.*

# 5. The measurement is stable across scoring runs

A measurement that agrees with the ground truth 67% of the time but gives different answers each time it is asked is measuring the instrument, not the construct. Reliability is the prerequisite for operational use. This section establishes that the predicted rating is near-deterministic across independent scoring runs.

We ran the prompt function three times on the same text data. Each run was independent: the model had no memory of prior predictions. Across the 11,020 sessions evaluated on all three runs, the results are striking.

*Table 5. Run-to-run stability across three independent scoring runs.*

| Metric | Value |
| --- | --- |
| Pairwise exact agreement | 87% |
| Pairwise within ±1 | 99.9% |
| Sessions identical across all three runs | 80.9% |
| Pairwise Pearson correlation | 0.992 |

Four out of five sessions received the same predicted score on every run. Nearly every remaining session varied by a single point. The total variation distance between run distributions is 0.005: effectively zero.

At the individual level, the consistency is tangible. The fan who wrote about parking costing forty dollars and vowed to "only attend if invited or win tickets" received a predicted score of 5 on all three independent runs. The model does not waver: given the same text, it reaches the same conclusion every time.

**This level of reproducibility means that the primary challenge is not measurement noise.** The model is consistent about what it sees in the text. When predictions diverge from fan scores, it is not because the model is guessing; it is because the model and the fan are responding to different signals in the same experience. This reframes the accuracy question: the gap is not stochastic error to be reduced through averaging. It is a systematic construct difference to be understood or corrected.

The scale of this consistency deserves context. In traditional content analysis, human intercoder reliability on subjective coding tasks typically falls in the range of Cohen's κ = 0.60–0.75: what the literature characterizes as "substantial" agreement (Krippendorff, 2018). The pairwise correlation of 0.992 and exact agreement rate of 87% reported here would be considered exceptional in terms of human intercoder reliability benchmarks. This consistency means that practitioners can treat the predicted rating as a stable baseline: any observed change in predicted scores over time reflects a real change in what fans are writing about, not drift in the measurement instrument. Further, it is evidence that the LLM is measuring something real and reproducible.

      This has two practical implications. First, a single scoring run is sufficient. Multiple runs add cost without meaningful information gain. Second, practitioners can treat the predicted rating as a stable baseline: any observed change in predicted scores over time reflects a real change in what fans are writing about, not drift in the measurement instrument.

## 6. The predicted rating tracks overall experience, not individual aspects

If the model were simply reacting to parking complaints or concession mentions, the predicted rating would correlate most strongly with those aspect ratings. It does not. The predicted rating's strongest relationship is with the overall experience score: not with any single aspect.

*Table 6. The predicted rating mirrors the correlation structure of the overall experience score rather than any individual aspect.*

| Variable | Corr. with predicted | Corr. with overall |
|---|---|---|
| overall_numrat | 0.816 | — |
| staff_numrat | 0.515 | 0.566 |
| concessions_numrat | 0.508 | 0.531 |
| entertainment_numrat | 0.454 | 0.502 |
| seatview_numrat | 0.429 | 0.490 |
| merchandise_numrat | 0.416 | 0.449 |
| parking_numrat | 0.363 | 0.389 |
| parking_exit_numrat | 0.287 | 0.282 |

The correlation between the predicted rating and the overall experience score (0.82) is substantially higher than the correlation between the predicted rating and any aspect (the highest is staff at 0.52). The predicted rating mirrors the correlation structure of the overall score itself: both are most correlated with staff and concessions, least correlated with parking exit. The model appears to be reading the same integrated signal that the overall score captures, not anchoring on any single dimension.

This integration is visible in individual cases. One fan wrote: "**I only rated a 10 because the staff was incredibly nice, the food was fantastic**… But Bob, sell the team." The text references staff, food, and team performance in a single sentence, weaving multiple dimensions into a holistic assessment. The model predicted 10, matching the fan's actual score and confirming that the prediction tracks the integrated verdict rather than any single complaint or compliment.

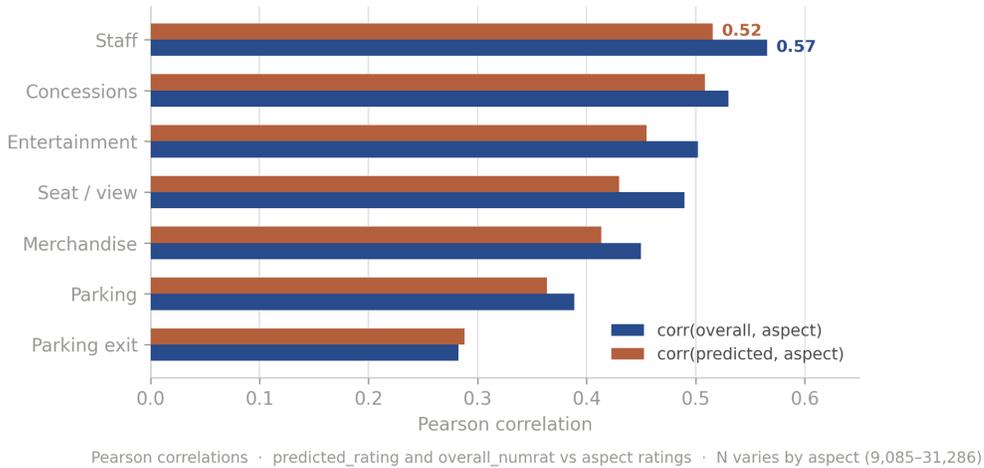

*Figure 4. The predicted rating mirrors the correlation structure of the overall score, not any single aspect. Both measures show the same rank ordering of aspects, with the predicted rating attenuated by approximately 0.05–0.10 across each.*

This finding recontextualizes the underprediction documented in Section 4. If the model were simply overweighting parking complaints, we would expect the delta between predicted and actual scores (predicted − overall) to correlate with parking ratings. It does not. The delta correlations with all aspect ratings are near zero (ranging from −0.03 to +0.05), indicating that the model's underprediction is broadly distributed across the experience rather than driven by any specific friction. The gap is not about parking. It is not about concessions. It is about something more general, a systematic difference in what the text emphasizes versus what the survey captures. Section 8 develops this interpretation.

# 7. Where and why predictions diverge

The specific patterns of divergence in individual cases reveal the structure of what text captures versus what surveys measure.

The most informative divergences are not random. They cluster into recognizable types. The most common and most consequential is what we term **fragile satisfaction**: fans who reported high overall scores (9 or 10) but whose text was dominated by a specific friction. Consider a fan who rated her experience a 10 but wrote about arriving two hours early for a promotional jersey and being turned away for having the wrong ticket type. The model read the frustration in the text and predicted a 3. The fan's verdict was about the day. The text was about the moment.

Other recurring patterns include **text-dominated friction** (a single vivid incident drives the prediction far below the overall score i.e., a security confrontation, an obstructed view), **sarcastic positivity** (a fan writes "I appreciated waiting in the long concession lines" and rates a 10; the model reads the surface complaint), and **reverse divergence** (fans who scored low but wrote positively about specific moments). Each of these patterns reflects a genuine discrepancy between what the fan chose to write about and the number they chose to give, not a failure of reading comprehension.

*Table 7. Representative verbatim examples illustrating each divergence pattern.*

| Pattern | Actual | Pred. | Fan text (excerpt) |
| --- | --- | --- | --- |
| Fragile satisfaction | 10 | 3 | Arrived two hours early for promotional jersey, turned away for wrong ticket type |
| Fragile satisfaction | 10 | 5 | "Everything was great... but parking... $40... rip off" |
| Text-dominated friction | 9 | 1 | Accused of stealing, threatened by security, "lost a fan for life" |
| Sarcastic positivity | 10 | 3 | "Appreciated waiting in the long concession lines. Loved the strain in my neck" |
| Reverse divergence | 1 | 9 | "We loved the game... awesome job" handling rain delay |
| Clean alignment | 10 | 10 | "Every team member pleasant, seats great, ballpark beautiful" |
| Clean alignment | 1 | 2 | "Such a disappointment... concessions 40 min... barely saw the game" |

Of these patterns, fragile satisfaction is the most consequential for practitioners. A fan base with high average self-reported ratings but a high incidence of fragile satisfaction (those rating 9 or 10 whose text is dominated by operational friction) is structurally vulnerable. Their satisfaction is real but conditional: it rests on affinity, habit, or a single peak moment rather than on consistently positive operations.

One additional friction, a price increase, a policy change, a bad game on a bad day, and the rating drops to match the text. Fragile satisfaction is invisible in aggregate rating dashboards. A team averaging 8.5 in overall experience looks healthy. But if 30% of those high-scoring fans wrote about concession delays, parking costs, or undelivered promotions, the 8.5 is fragile. The predicted rating surfaces this vulnerability because it measures what the fan chose to report, not what the fan concluded. Tracking the gap between predicted and self-reported scores over time creates an early warning system: a widening gap means fans are writing about more friction while still evaluating the day positively. That gap will eventually close: either because operations improve (the text catches up to the verdict) or because satisfaction erodes (the verdict catches up to the text).

These examples are not cherry-picked outliers. They represent systematic patterns visible across the dataset, and they share a common structure: the text and the verdict are not reporting on the same thing. Section 8 develops the account of the construct difference the gap reflects.

# 8. What the divergence reveals: salience versus verdict

The systematic underprediction documented in Sections 4 and 7 is not measurement error in the conventional sense. It is a construct difference. When a fan writes about their experience, they write about what was most **salient**: the moments that were most memorable, emotionally intense, unusual, or actionable. When the same fan answers a survey question, they render a **verdict**: an explicit overall judgment that integrates their entire day, including the parts they did not bother to mention.

One fan made both constructs visible in a single sentence: "A 10 but rating this exact game a 2. We lost game 1 to the A's!" The 10 was the verdict, her overall relationship with the experience of attending. The 2 was the salient reaction to what happened that day. The model read the text, saw the loss and the explicit 2, and predicted accordingly. It was not wrong. It was measuring a different thing. This case is extreme, but the mechanism reveals a fan whose overall judgment and whose written account points in different directions, operates throughout the dataset, most visibly in the 7–8 band where fans had a good day but wrote about what went wrong.

## Two constructs, one fan: verdict versus salience
*The diagonal is where the model works; the off-diagonal is where the construct difference lives*

|  | POSITIVE TEXT | NEGATIVE TEXT |
|---|---|---|
| **High** | **Clean alignment**<br>"*Every team member encountered was pleasant and interactive, seats were great, ballpark is beautiful*"<br>Actual 10 · Predicted 10<br>*Text and verdict agree* | **Fragile satisfaction**<br>"*Tail-gating awesome… Stadium atmosphere awesome. Team and Manager SUCKS… DUMB organization…*"<br>Actual 10 · Predicted 4<br>*Text salience pulls prediction down* |
| **Low** | **Reverse divergence**<br>"*We loved the game… awesome job*" handling rain delay<br>Actual 1 · Predicted 9<br>*Positive language masks a harsh verdict* | **Clean alignment**<br>"*Such a disappointment… concessions 40 minutes… barely saw any of the game*"<br>Actual 1 · Predicted 2<br>*Text and verdict agree* |

Text sentiment →

Representative verbatims by self-reported score (high/low) and text sentiment (positive/negative) · MLB fan experience surveys, Mar–Jun 2025

*Figure 5. Two constructs, one fan: representative verbatims arranged by self-reported score (high/low) and text sentiment (positive/negative). The diagonal cells are where text and verdict agree and the model works. The off-diagonal cells are where the construct difference lives.*

This distinction between salience and verdict explains the empirical patterns observed throughout this study. The underprediction is consistent and directional (fans write about frictions; the model reads the frictions and lowers its estimate). It concentrates in the 7–8 band (fans with good experiences who wrote about what went wrong). It is broadly distributed across aspects rather than driven by any single dimension. And it is strongest in medium-confidence cases (where the text contains mixed signals rather than a clear verdict).

The cognitive science literature provides a framework for understanding why fans' written and numeric responses diverge in this particular direction and the fact that both questions appear on the same survey sharpens rather than undermines the account. Because the open-ended response and the rating scale are completed in immediate sequence, the divergence cannot be attributed to differential memory decay or context shift. The same game-day experience is equally accessible when the fan answers both questions. What differs is the cognitive operation each format invites.

Schwarz (1999) demonstrates that open-ended prompts activate specific incident reporting: respondents retrieve concrete, vivid, and easily describable moments and put them into words. Rating scales activate a different process — holistic evaluative judgment that draws on global affective summary. Schwarz and Clore (1983) formalize this as the mood-as-information model: when asked for an overall evaluation, respondents consult their current affective state as a primary input rather than summing individual episodes. The result is that routine positive elements of the experience (a comfortable seat, an exciting game, a cold beer) contribute to the affective state that anchors the rating but rarely generate the kind of specific, reportable content that appears in open-ended text. Negative and surprising events, by contrast, tend to be more report-worthy: they are more informative, less expected, and more actionable for the reader (Baumeister et al., 2001). A fan who writes about long concession lines and expensive parking is not necessarily describing a bad day: she is describing what, under Gricean norms of communication, is worth saying.

The survey question, by contrast, asks the fan to step back and render a verdict on the day as a whole. This global evaluation integrates across the entire experience, including the routine satisfaction that never surfaces in writing. Schwarz and Strack (1999) describe this as a distinction between episodic retrieval (specific moments) and evaluative judgment (overall assessment). The open-ended response captures the former; the survey score captures the latter.

But question format is not the only asymmetry. The verdict also draws on inputs that are dispositional rather than episodic: the fan's relationship with the team, the identity that comes with being a season-ticket holder or a lifelong follower, the accumulated affinity that makes attending a game worth doing even when nothing remarkable happens. These inputs raise the baseline of the overall rating without generating reportable content, because they are not things that happened today. No fan writes "the stadium felt like home and I've loved this team since I was eight" in a post-game survey, but that sentiment is present in the number she gives. The fan who wrote "A 10 but rating this exact game a 2" made the structure visible: the 10 integrates a relationship with the experience of attending that predates and outlasts any single game; the 2 is about what happened on that Tuesday. Open-ended text is often exclusively episodic while the verdict more often contains a mixture.

This framing has a practical consequence. The gap between predicted and actual scores is not noise to be eliminated. It is a signal to be preserved. The fan who wrote "A 10 but rating this exact game a 2. We lost game 1 to the A's!" expressed both numbers explicitly. The 10 and the 2 are not contradictions; they are two true and compatible reports about two different things. The predicted rating tracked the salient reaction to what happened that day; the survey score captured the broader verdict about the experience of attending. Each contains information the other misses. A team that sees only the survey score will miss the operational friction. A team that sees only the predicted rating will underestimate fan satisfaction.

Two counterarguments deserve acknowledgment. First, the underprediction could partly reflect **negativity bias in open-ended responses**: fans may be more likely to write at all when they have a complaint. This study cannot separate sample selection effects from within-respondent salience effects because all fans in the sample both wrote and scored. Second, the model itself may **overweight negative language**: words like "awful," "rip off," and "ruined" may carry disproportionate weight in the model's inference relative to their weight in the fan's overall assessment. The impact of revisions to the prompt instructions should be explored in future research.

What this study can establish is that the gap is real, systematic, stable, and not reducible to a single aspect of the experience. It is a construct difference, not an instrument failure.

## 9. Conclusions

Predicted ratings derived from fan text are a reliable, stable indicator of fan-reported overall experience. Across approximately 10,000 sessions and three independent scoring runs using GPT-4.1, 67% of predicted ratings fell within one point of self-reported scores, and the measurement was near-deterministic in its reproducibility (87% exact agreement across runs, 99.9% within ±1). These results are considered excellent in terms of human inter-coder reliability. The predicted rating tracks overall experience ($r = 0.82$) rather than any individual aspect of the game-day experience, and it captures daily trends in fan experience with a correlation of 0.92.

**Is perfect alignment the correct goal?** This study validated predicted ratings against fan-reported scores, and practitioners may naturally ask how to close the remaining gap. But the more important question is whether closing it is the right objective. Perfect alignment would mean the predicted rating reproduces the survey score exactly. However, the deeper value of the predicted rating lies precisely in what it captures that the survey score does not.

A fan's overall experience rating is a verdict, a single number that compresses an entire game day into one point on an 11-point scale. Two fans can both rate a 10 for entirely different reasons, and two fans can both rate a 10 while harboring entirely different operational frictions. The number tells you how the fan evaluated the day. It does not tell you what shaped that evaluation, and it does not tell you what to do differently. The open-ended response preserves the specifics the rating discards: the concession line that took forty minutes, the parking attendant who yelled, the promotional jersey that was promised and not delivered. These are not summaries of the overall experience. Rather, they are the moments the fan selected as worth reporting and that selection is itself informative. Fredrickson and Kahneman's (1993) peak-end rule offers a partial account of why the survey score is not a simple average of the experience: retrospective evaluations are dominated by the emotional peak and the final moment, not by an integration of every episode. This is one reason why two fans can rate a 10 for entirely different reasons — what the number reflects is the shape of the experience, not its content. Kahneman and Riis (2005) formalize the broader distinction as the difference between the experiencing self, which lives through every moment, and the remembering self, which retains a handful of representative moments and uses them to construct a verdict. Both the open-ended response and the survey score are products of the remembering self and both are completed after the experience ends. What differs is the cognitive operation each format invites: the rating scale asks for a summary verdict, and the open-ended prompt asks for reportable specifics, producing the construct divergence this paper documents.

The strongest analytical position is not to choose between the two but to use both. The self-reported score tells you the fan's verdict: how they weighed the full day when asked to put a number on it. The predicted score tells you what the text emphasized: the moments most cognitively available when the fan sat down to write. When the two agree, confidence in both signals is high. When they diverge (i.e., a fan who rates a 10 but whose text produces a predicted 5) the divergence itself is the finding. It identifies fans whose satisfaction is real but whose experience contains operational friction worth investigating. Neither score alone surfaces this pattern. A dashboard of average self-reported ratings can tell you that satisfaction dipped in June. It cannot tell you that the dip was driven by a new parking policy or a promotion that ran out of inventory. A dashboard of predicted ratings alone will underestimate how satisfied fans actually are. Together, the two scores answer different questions: the self-reported rating answers "how are we doing?" and the predicted rating answers "what are fans talking about, and how does it feel?"

This framing also clarifies the value of predicted scores in contexts where no self-reported rating exists for comparison. Most textual feedback in operational settings arrives without an accompanying survey score: support transcripts, social media posts, online reviews, unsolicited emails, in-app feedback. In these contexts, there is no ground truth to validate against, and the question of alignment is moot. What remains is a text and a model that can read it. The predicted score converts unstructured language into a structured, comparable signal: a number that can be tracked over time, segmented by venue or product, and used to identify trends that would otherwise require a human analyst to read thousands of verbatims.

Finally, the stability results in this study (Section 5) are directly relevant here: because the measurement is near-deterministic, practitioners can trust that changes in predicted scores over time reflect real changes in what people are writing about, not noise in the instrument. The predicted rating does not need a survey score to be useful. It needs only text and a well-defined construct. The absolute values of predicted scores should, however, be interpreted with caution given known properties of LLM scalar measurement (compression, heaping), and the predicted score is most trustworthy as a ranking and trend-tracking instrument rather than as an absolute measure (Licht et al., 2025). The validation in this paper establishes that when a survey score does exist, the two are meaningfully related and extremely reliable. This gives practitioners reason to trust the predicted score in the much larger universe of text where no survey score is available.

# 10. Implications for data practitioners

**These are baseline results. Treat them as a floor, not a ceiling.** The predicted rating was generated by a simple, unoptimized prompt with no examples, no calibration, and no domain-specific engineering. Any operational deployment would begin here.

**A single scoring run is sufficient.** The near-deterministic stability across three independent runs (87% exact agreement, 99.9% within ±1, r = 0.992) means that running the model multiple times adds cost without meaningful information gain. If the predicted rating changes between two time periods, the change reflects a real shift in what fans are writing about.

**Use the confidence label as an operational filter.** Restricting to High-confidence predictions raises within ±1 agreement from 67% to 82% and roughly doubles exact-match accuracy. For applications requiring high fidelity, filtering by confidence is the highest-leverage intervention available without changing the prompt. More broadly, aggregate metrics blend High and Medium confidence predictions, which behave very differently. Build this distinction into dashboards and reporting from the start.

**Do not optimize for perfect alignment with survey scores.** The gap between predicted and self-reported ratings is not error to be eliminated. It is information to be used. The predicted rating captures what fans chose to write about, the operationally salient moments, while the survey score captures their overall verdict.

For organizations managing experience at scale, the rating answers "how are we doing?" while the text answers "what should we fix?" and the gap between predicted and self-reported scores is itself an early warning signal that aggregate satisfaction metrics cannot provide.

**Use the divergence as a diagnostic.** When the average predicted score drops but the average self-reported score holds steady, fans are writing about more friction but still evaluating the day positively. This pattern of "fragile satisfaction" is an early warning signal that aggregate satisfaction metrics will not detect. When the two scores move together, the signal is unambiguous.

**Start with calibration, not prompt revision.** Before modifying the prompt, apply a calibration mapping to see how much of the gap it closes. A simple lookup from predicted score to expected actual score will align the measurement with survey conventions without changing what the model measures. Even without calibration, the predicted rating is most reliable as a ranking and trend-tracking tool rather than an absolute threshold: the compressed scale means a predicted 7 corresponds to a fan-reported 8+.

# 11. Limitations

**Sample composition uncertainty.**  The dataset was filtered for test traffic and non-fan responses. As such the alignment metrics reported here represent a strong base for comparing accuracy against genuine fan responses.

**Single context.**  This study examines fan experience surveys from five MLB teams. Whether the same alignment levels hold in other domains is an empirical question this study cannot answer.

**No multivariate modeling.**  We report bivariate correlations between predicted scores and aspect ratings. A multivariate regression may help to identify which aspects independently explain the gap.

**No gold standard exists for this task.** We validate predicted ratings against self-reported survey scores because the survey score is the operationally relevant reference point, i.e., what the fan actually reported. We do not assess agreement between the model and human coders on this task, but we would expect human coders to also diverge from the fan's rating, for the same cognitive reasons documented in Section 8: the text emphasizes salient friction while the rating integrates the full experience including routine positives that never surface in writing. The absence of a gold standard does not invalidate the measurement. It means that validation must rest on reliability (Section 5), construct validity (Section 6), and interpretive coherence (Section 8) rather than on agreement with any single reference.

**Single model.**  All annotation was conducted with GPT-4.1. Replication with alternative models would strengthen the generalizability of these baseline results.

## Appendix A: Prompt function schema

The analysis in this study was conducted using the Dimension Labs language data platform ([dimensionless.io](dimensionless.io)). The complete JSON schema used to generate predicted ratings is reproduced below. See Section 3.2 for discussion.

```
{
  "name": "extract_session_info",
  "description": "Infer the fan's self-reported overall game-day experience rating from their open-ended feedback.",
  "context": "Incoming messages are from the fan. Outgoing messages are survey questions.",
  "parameters": {
    "type": "object",
    "properties": {
      "predicted_rating": {
        "description": "Estimate the fan's self-reported overall experience rating (0-10, 10 best) as they would answer the survey question 'overall experience attending the game'. Use ONLY evidence in the fan's text. If there is not enough evidence to infer an overall rating with confidence, output null.",
        "type": "integer"
      },
      "predicted_rating_confidence_score": {
        "description": "Confidence in the inferred overall rating based on clarity and strength of evidence in the fan text. High: explicit rating or very clear overall sentiment with strong cues. Medium: inferable but mixed/limited cues. Low: weak cues or ambiguous. If predicted_rating is null, output null.",
        "type": "string",
        "enum": ["High", "Medium", "Low"]
      },
      "predicted_rating_evidence": {
        "description": "In <15 words, summarize the strongest evidence from the fan text for the inferred overall rating. all lower case, no punctuation. If predicted_rating is null, output null.",
        "type": "string"
      }
    },
    "required": ["predicted_rating", "predicted_rating_confidence_score", "predicted_rating_evidence"]
  }
}
```

# Appendix B: Supplementary tables

*Table B1. Full score-level alignment metrics (actual 0–10, run p2_1).*

| Actual | N | Exact | Within ±1 | MAE | Bias |
|---|---|---|---|---|---|
| 0 | 234 | 12.8% | 34.2% | 1.86 | +1.86 |
| 1 | 185 | 17.8% | 68.1% | 1.34 | +1.34 |
| 2 | 249 | 36.1% | 82.7% | 0.89 | +0.36 |
| 3 | 268 | 39.6% | 85.1% | 0.81 | −0.07 |
| 4 | 267 | 24.0% | 66.3% | 1.17 | −0.77 |
| 5 | 603 | 12.4% | 39.6% | 1.72 | −1.43 |
| 6 | 540 | 12.4% | 34.4% | 2.11 | −1.96 |
| 7 | 1,144 | 23.3% | 45.5% | 2.01 | −1.82 |
| 8 | 1,751 | 26.8% | 57.7% | 1.79 | −1.54 |
| 9 | 1,547 | 29.9% | 69.3% | 1.44 | −1.19 |
| 10 | 3,187 | 61.7% | 89.5% | 0.52 | −0.48 |

# Appendix C: Verbatim reference examples by divergence pattern

A broader set of examples organized by divergence pattern, for analyst reference.

### D1. Fragile satisfaction (actual high, predicted low)

| Actual | Pred. | Conf. | Fan text (excerpt) |
|---|---|---|---|
| 10 | 2 | High | "A 10 but rating this exact game a 2. We lost game 1 to the A's!" |
| 10 | 3 | High | Arrived first in line, waited 2 hours for promotional jersey, turned away |
| 10 | 3 | Med. | "I appreciated waiting in the long concession lines. Loved the strain in my neck" |
| 10 | 4 | Med. | "Tail-gating awesome... Stadium atmosphere awesome. Team and Manager SUCKS" |
| 10 | 5 | Med. | "Everything was great... But parking... $40... rip off" |

### D2. Text-dominated friction (actual high, predicted very low)

| Actual | Pred. | Conf. | Fan text (excerpt) |
|---|---|---|---|
| 9 | 1 | High | "Accused of stealing... threatened by security... lost a fan for life" |
| 9 | 2 | High | "Female parking attendant yelled... great customer service" (sarcastic) |
| 9 | 2 | High | "Obstructed view... not able to see either the pitcher or batter" |

### D3. Reverse divergence (actual low, predicted high)

| Actual | Pred. | Conf. | Fan text (excerpt) |
|---|---|---|---|
| 1 | 9 | High | "We loved the game... awesome job" handling rain delay; no follow-up on tickets |
| 1 | 7 | Med. | "Only 20,000 bobbleheads... didn't get one... I understand" |

### D4. Clean alignment (|predicted − actual| ≤ 1)

| Actual | Pred. | Conf. | Fan text (excerpt) |
|---|---|---|---|
| 10 | 10 | High | "I only rated a 10 because the staff was incredibly nice, the food was fantastic" |
| 10 | 10 | High | "Every team member pleasant, seats great, ballpark beautiful" |
| 1 | 2 | High | "Such a disappointment... concessions 40 minutes... barely saw any of the game" |
| 10 | 10 | High | "11/10 amazing experience" |